\documentclass[10pt, a4paper]{article}

\usepackage[final]{lrec2026}

\usepackage{graphicx}

\usepackage{subcaption}

\usepackage[linesnumbered,ruled,vlined]{algorithm2e}
\SetKwInput{KwInit}{Initialize}
\SetKwInput{KwParams}{Parameters}

\usepackage{tcolorbox}
\usepackage{cleveref}
\crefname{subsection}{subsection}{subsections}
\crefname{algorithm}{algorithm}{algorithms}
\Crefname{algorithm}{Algorithm}{Algorithms}
\usepackage{minted}
\usepackage{booktabs}

\newcommand{\fon}[1]{\fontfamily{#1}\selectfont}

\title{\textsc{DeepQuestion}: Systematic Generation of Real-World Challenges for Evaluating LLMs Performance}

\name{Ali Khoramfar, Ali Ramezani, Mohammad Mahdi Mohajeri \\
{\bf \large Mohammad Javad Dousti, Majid Nili Ahmadabadi, Heshaam Faili} }

\address{Department of Electrical and Computer Engineering, College of Engineering, \\
University of Tehran, Tehran, Iran \\
\{khoramfar, ali.ramezani.96, mehdimohajeri, mjdousti, mnili, hfaili\}@ut.ac.ir}

\abstract{
While Large Language Models (LLMs) achieve near-human performance on standard benchmarks, their capabilities often fail to generalize to complex, real-world problems. To bridge this gap, we introduce \textsc{DeepQuestion}, a scalable, automated framework that systematically elevates the cognitive complexity of existing datasets through controlled task transformations grounded in explicit cognitive hierarchies. Based on Bloom's taxonomy, \textsc{DeepQuestion} generates (1) scenario-based problems to test the application of knowledge in noisy, realistic contexts, and (2) instruction-based prompts that require models to create new questions from a given solution path, assessing synthesis and evaluation skills. Our extensive evaluation across ten leading open-source and proprietary models, covering both general-purpose and reasoning LLMs, reveals a stark performance decline—with accuracy dropping by up to 70\%—as tasks ascend the cognitive hierarchy across evaluation settings. These findings underscore that current benchmarks overestimate true reasoning abilities and highlight the critical need for cognitively diverse evaluations to guide future LLM development.
 \\ \newline 
}
 
\begin{document}
\maketitleabstract

\section{Introduction}
Recent advances in Large Language Models (LLMs) have driven remarkable improvements across a wide spectrum of benchmarks, from arithmetic reasoning in GSM8K to expert-level performance on MMLU and GPQA~\cite{achiam2023GPT4, team2023gemini, grattafiori2024llama3, team2025gemma3, cobbe2021gsm8k, hendrycks2math, hendryckstest2021mmlu, wang2024mmlupro, rein2024gpqa}. Yet, as scores approach saturation, evidence increasingly suggests that such benchmarks do not reflect the demands of real-world reasoning. When confronted with tasks that include irrelevant details, ambiguous contexts, or creative synthesis, even state-of-the-art LLMs exhibit significant failures—revealing an overreliance on pattern recognition rather than genuine understanding~\cite{arias2025automatic, myrzakhan2024open}.

Emerging studies have begun to expose this gap. Models that perform flawlessly on structured academic datasets often falter on problems drawn from authentic or modern settings, such as new mathematics contests, clinical scenarios, or symbolic reasoning variations~\cite{shojaee2025illusion, mirzadehgsm, petrov2025proof, alaa2025medical}. These shortcomings highlight that current benchmarks primarily target lower-order cognitive abilities—such as recall and comprehension—while neglecting the deeper reasoning, analysis, and creativity required for practical decision-making. Without cognitively diverse evaluation frameworks, progress risks being misinterpreted as intelligence rather than memorization.





\begin{figure}[t]
    \centering
    \includegraphics[ width=0.60\columnwidth]{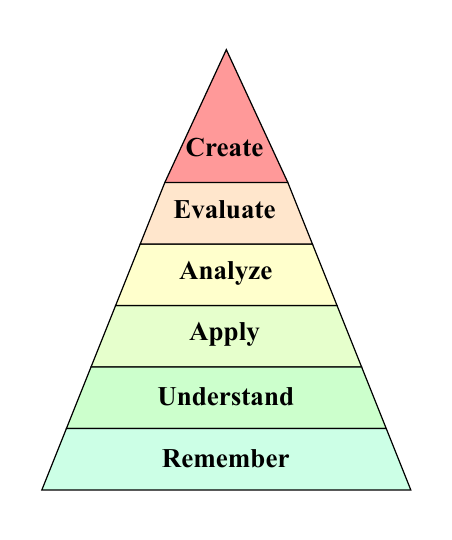}
    \caption{Bloom's taxonomy hierarchy.}
    \label{fig:bloom}
\end{figure}

To address this limitation, we introduce \mbox{\textsc{DeepQuestion}}, a systematic framework for generating cognitively enriched evaluation datasets grounded in Bloom’s taxonomy of learning~\cite{forehand2010bloom,leung2000assessment}, a widely recognized framework outlining six cognitive skill levels, from basic recall to higher-order reasoning (see \Cref{fig:bloom}). \textsc{DeepQuestion} transforms existing question–answer pairs into (1) scenario-based problems that test the application of knowledge in realistic, constraint-rich situations, and (2) instruction-based prompts that assess creation and evaluation skills by requiring models to design new questions aligned with given solution paths. This taxonomy-guided approach adds interpretable layers of complexity that correspond directly to human cognitive processes.

%
%
Through extensive experiments across ten leading LLMs—spanning general-purpose and reasoning-oriented architectures—we observe sharp performance declines, up to 70\%, as the cognitive level of tasks increases. These results provide evidence that LLMs’ strong benchmark performance does not generalize to deeper reasoning. Beyond evaluation, the \textsc{DeepQuestion} framework offers a replicable foundation for constructing cognitively meaningful benchmarks across diverse disciplines, from mathematics and physics to law and medicine.
Our main contributions are:
\begin{itemize}
    \item Proposed \textsc{DeepQuestion}, a framework for generating questions from existing datasets for better LLM evaluation based on Bloom's taxonomy.
    \item Introduced the \textsc{DeepQuestion} dataset, created using our framework.
    \item Conducted a comprehensive evaluation of LLMs with the \textsc{DeepQuestion} dataset, highlighting their knowledge limitations across Bloom's taxonomy.
\end{itemize}

The remainder of this paper is organized as follows: \Cref{sec:related} reviews the related works, \Cref{sec:method} introduces the \textsc{DeepQuestion} framework, \Cref{sec:setup} details our experimental setup, and the final sections present our results and conclusions.

\begin{figure*}[t]
    \centering
    \includegraphics[ width=1\textwidth]{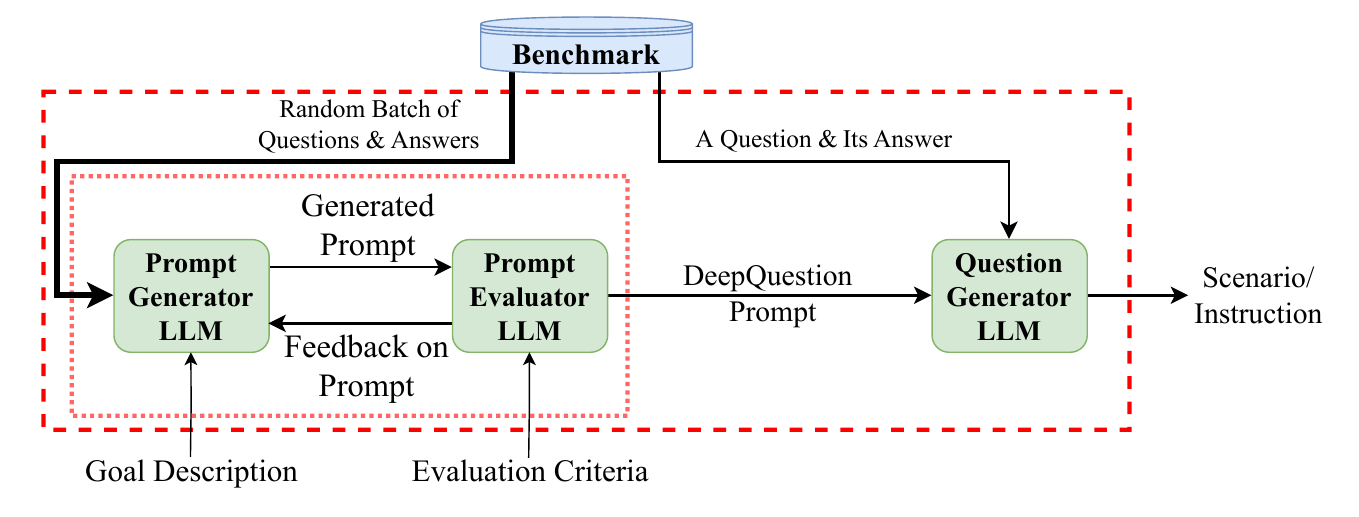}
    \caption{Overview of \textsc{DeepQuestion} framework. It begins with the selection of a random batch of questions and answers. Then, by conversation between the prompt generator and the prompt evaluator LLMs, the deep-question prompt is generated. The question generator LLM with the deep-question prompt converts each question and answer pair to the deep question and answer.}
    \label{fig:Framework}
\end{figure*}

\section{Related Works}
\label{sec:related}
This paper draws on ideas from two main areas: the levels of learning, derived from the learning sciences, and the limitations of LLMs in solving complex and real-world problems. In this section, we review the previous work related to these two domains.

\subsection{Learning Levels}
In the learning sciences, various definitions of learning have been proposed. Some of these frameworks conceptualize learning as occurring at different levels. Among the most influential are the SOLO taxonomy and Bloom’s taxonomy, both of which define learning across hierarchical stages~\cite{ilhan2017comparison}.

In the SOLO taxonomy~\cite{biggs2014evaluating}, the lowest level of learning is Pre-structural, where no meaningful learning has occurred. The highest level, known as Extended Abstract, indicates that the learner is able to extend their understanding and apply their knowledge to new domains.

Bloom’s taxonomy~\cite{bloom1971taxonomy, anderson2001taxonomy, forehand2010bloom} offers a more widely adopted framework. It also defines learning in multiple levels and exists in several versions. In its most recent revision, the lowest level is \emph{Remember}, in which the learner simply recalls information without deeper understanding. A higher level, apply (the third level), represents the learner’s ability to use acquired knowledge to solve real-world problems. The highest level, \emph{Create}, reflects the learner’s capacity to generate new ideas or concepts based on previously acquired knowledge.

Employing these taxonomies to assess the performance of LLMs can provide a more human-aligned perspective on evaluating their knowledge and reasoning capabilities.

\subsection{LLMs Evaluation in Complex and Real-world Problems}
With the rapid advancement of LLMs, their ability to solve complex problems has become one of the key questions in the field. Moreover, as the potential use of LLMs in real-world contexts grows, evaluating their performance on real-world problems has become increasingly important.

MMLU~\cite{hendryckstest2021mmlu} serves as a standard benchmark for evaluating LLMs across diverse academic and professional domains. However, as model performance on MMLU has approached saturation, MMLU-Pro~\cite{wang2024mmlupro} was introduced to assess models on more complex, reasoning-intensive tasks. Establishing such advanced benchmarks is essential to accurately measure progress and ensure meaningful evaluation of LLMs’ capabilities in challenging real-world scenarios. This aligns with findings from The Illusion of Thinking~\cite{shojaee2025illusion}, which highlight the limitations of current LLMs when confronted with more complex reasoning tasks.

Recent studies~\cite{chauhan2025assessing, alaa2025medical} have examined the performance of LLMs in real-world scenarios and compared it with their performance on standard benchmarks. These studies indicate a notable reduction in LLM performance when applied to real-world contexts. Given the growing interest in deploying these models in practical applications, assessing their performance in real-world settings is crucial. Consequently, rigorous evaluation of LLMs under such conditions has become increasingly important.

\begin{figure*}[t]
    \centering

    \includegraphics[ width=1\textwidth]{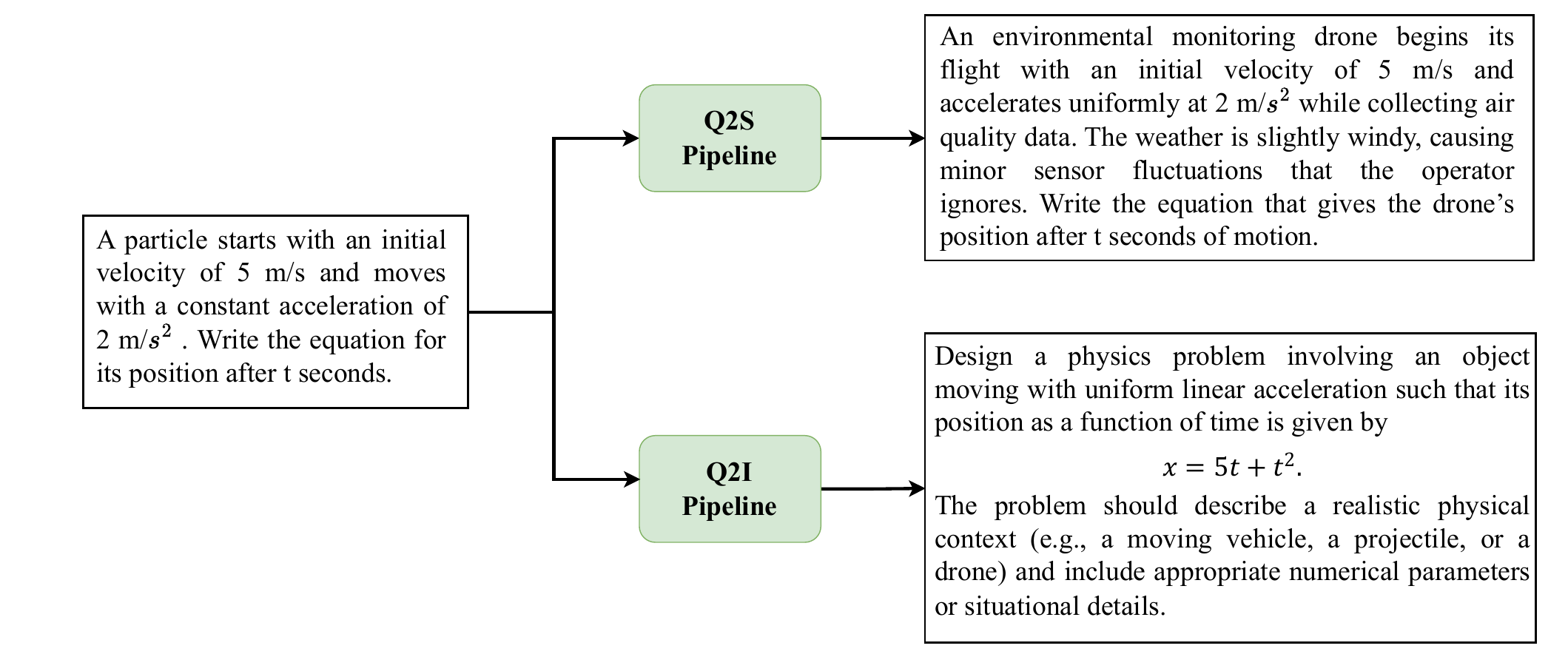}
    \caption{Examples of question transformations produced by the Q2S and Q2I pipelines}
    \label{fig:examples}
\end{figure*}

\section{Method}
\label{sec:method}
This study is guided by a key research question: Can existing benchmarks be systematically deepened to reflect real-world task complexity, rather than introducing artificial difficulty?

To address this, we leverage insights from learning science—particularly Bloom’s taxonomy—which offers a principled framework for systematically controlling and measuring cognitive complexity. Unlike ad-hoc increases in difficulty, Bloom’s hierarchy allows us to add real, interpretable complexity to benchmark items, specifying exactly how and by how much each transformation increases the cognitive demands.

Prior analyses reveal that most benchmarks remain at lower Bloom levels—\emph{Remember} and \emph{Understand}~\cite{alaa2025medical, chauhan2025assessing}—and rarely stress the higher-order reasoning required in practical tasks. To bridge this gap, we focus on the \emph{apply} and \emph{create} levels, which represent critical forms of real-world reasoning often underrepresented in evaluation. Specifically, we propose two aligned methods: (1) question-to-scenario (Q2S), which increases complexity by embedding items in realistic, constraint-rich contexts to elicit \emph{apply}-level reasoning; and (2) question-to-instruction (Q2I), which reframes items as multi-step, constructive procedures with explicit acceptance criteria, targeting the \emph{create} level of synthesis and innovation. These transformations go beyond surface-level difficulty and require test-takers to transfer, adapt, and synthesize knowledge in meaningful, task-driven ways.

By systematically deepening benchmarks along Bloom’s hierarchy using Q2S and Q2I, our \textsc{DeepQuestion} framework enables precise diagnosis of model weaknesses—revealing, for example, whether a model fails at context adaptation, constraint handling, or constructive synthesis. This fine-grained understanding directly informs concrete intervention strategies for model improvement, such as targeted data augmentation, objective design, or architectural changes. Implementation details and our automated prompt-generation pipeline are described in the following section.

\subsection{Question-to-Scenario Generation (Q2S)}
\label{sec:q2s}
To deepen questions, Q2S targets the third level of Bloom’s taxonomy, namely \emph{Apply}, which emphasizes using knowledge in practical contexts. Each original question is transformed into a scenario-based version embedding the core problem within a realistic narrative, often including extraneous details and some distractions to simulate real-world complexity.
For example, as shown in \Cref{fig:examples}, a basic physics question asking for the position-time equation of an object, is reframed within a scenario involving an environmental drone, where irrelevant elements like weather acted as distractors. The core question data and required reasoning remain unchanged.
Q2S employs an automated prompt-based pipeline powered by an LLM (see \Cref{sec:pipe} for more details on the pipeline). The output is a benchmark of scenario-driven questions that better reflect authentic problem-solving contexts.

These scenario-based questions resemble real-world problems.
To solve them, LLMs must extract relevant information from the scenario and apply their knowledge accordingly.
This process reflects a higher cognitive level than the \emph{Remember} and \emph{Understand} levels.

\subsection{Question-to-Instruction Generation (Q2I)}
\label{sec:q2i}
Encouraged by the initial result analysis at the \emph{Apply} level of Bloom’s taxonomy with Q2S, we extend our approach to the higher \emph{Evaluate} and \emph{Create} levels.
At these levels, learners are expected to assess information critically and produce novel outputs.
We hypothesize that individuals or models with deep conceptual understanding are not only able to solve problems but also to design meaningful questions.
Designing a question requires the evaluation of existing knowledge in order to shape a coherent concept.
It also involves the ability to establish a logical and structured process that leads to the formulation of a meaningful question.
This process demands both critical reflection and creative synthesis of ideas.
Consequently, the capacity to design questions can be used to assess competencies at both the evaluate and create levels of cognition.
This informs our second experimental setup, which reverses the direction of the task: instead of solving a question, the model is instructed to design one.

For each original dataset question, Q2I constructs an instruction designed to prompt an LLM to generate a new question that preserves the same topic and solution path as the original. This instruction is formulated in a way that ensures a different LLM, when given the instruction, would also produce a question aligned with the original in both topic and reasoning process.
For example, if the original question in \Cref{fig:examples} (position-time equation of an object) yields the solution $x = 5t + t^2$, the instruction would ask the model to design a physics problem whose solution would be exactly that equation.
Designing questions based on instruction implicitly tests three levels of the model’s reasoning: conceptual understanding of the domain and relevant equations, the ability to define appropriate variables and their interrelations, and the selection of values that yield the target solution.

Q2I employs an automated prompt-based pipeline similar to Q2S; however, the output of Q2I is the set of instructions for question generation. The next subsection details the pipeline.

\subsection{Prompt Generation Pipeline}
\label{sec:pipe}
Since transforming each benchmark into its Q2S and Q2I variants requires carefully crafted prompts tailored to the specific domain and style of that benchmark, a key component of the \textsc{DeepQuestion} framework is a prompt generation pipeline which automates the creation of these task-specific prompts and is shown in \Cref{fig:Framework}. This pipeline eliminates the dependency on human expertise for prompt design and enables scalable adaptation to different benchmarks.

The pipeline operates through an iterative dialogue between two LLMs: a prompt generator and a prompt evaluator. Initially, a randomly selected batch of questions and their answers are sampled from the source benchmark, along with a high-level \textit{goal description}, are provided to the prompt generator LLM. The goal is to produce a prompt that can transform similar input questions into scenario-based questions in Q2S or instruction prompts in Q2I.

The generated prompt is then passed to the evaluator LLM, which assesses it based on predefined \textit{evaluation criteria}. The evaluator assigns a numerical score ranging from 0 to 10 and provides qualitative feedback describing strengths and weaknesses of the prompt.

If the score surpasses a threshold (e.g., 8), the prompt is accepted for use in the corresponding generation task. Otherwise, the prompt generator revises the prompt using the evaluator’s feedback, and the process repeats iteratively until a satisfactory prompt is produced. Once a prompt is accepted, it is then reviewed by a human domain expert, who either approves or rejects the generated prompts. In our experiments on the GSM8K and physics question datasets, the produced prompts were approved by the experts. See the generated prompt for Q2S for physics questions in the following.

This automatic prompt optimization is described in \Cref{algo}. By leveraging this pipeline, the \textsc{DeepQuestion} framework achieves robust prompt design tailored to the specific domain and style of each benchmark, enabling effective question transformation without manual intervention.

\begin{algorithm}
\caption{\textsc{DeepQuestion} Framework Algorithm}
\label{algo}
\KwIn{QA dataset $B = \{(q_i, a_i)\}$}
\KwOut{Deeper Question Benchmark: \textsc{DeepQuestion}}

$score \gets 0$;

$feedback \gets \texttt{None}$;

\texttt{goal\_description} $\gets$ Description of the intended purpose of generated prompts;

\texttt{evaluation\_criteria} $\gets$ Instructions for how the LLM should assess generated prompt quality;

Sample a random subset $B_s \subset B$;

\While{$score < 8$}{
    $prompt \gets \texttt{LLM}(goal\_description, B_s, feedback)$;
    
    $(score, feedback) \gets \texttt{LLM}(evaluation\_criteria, prompt)$;
}

\texttt{DeepQuestion} $\gets$ empty list;

\ForEach{$(q, a) \in B$}{
    $q' \gets \texttt{LLM}(prompt, q, a)$;
    \texttt{DeepQuestion.append}$(q', a)$;
}
\end{algorithm}

\begin{tcolorbox}[
fonttitle=\small\fon{pbk}\bfseries,
    left=2pt,
    right=2pt,
    top=2pt,
    bottom=2pt,
    title=Deep-Question Prompt Q2S for Physics Questions,
]
My goal is to design questions based on Bloom's *Apply* level. I will give you a question, and your task is to create a scenario in the form of a narrative that includes redundant material not related to solving the question. Ultimately, the explicit question I provide must be solved to find the answer, but the student must infer this question from the narrative.

The narrative should include a real-world story and irrelevant numbers that are ultimately simplified to the original problem.

* You should never give any hints about which information is necessary and which is irrelevant.
* The story should not reference the formulas or concepts required to solve the question. It should put the solver in a situation where they must apply their knowledge independently.
* All formulas and numbers should be presented in LaTeX format.
* All your answers must be in Persian.
\label{fig:prompt}

\end{tcolorbox}

\begin{figure*}[htbp]
    \centering
    \begin{subfigure}[b]{0.46\textwidth}
        \centering
        \includegraphics[width=\textwidth]{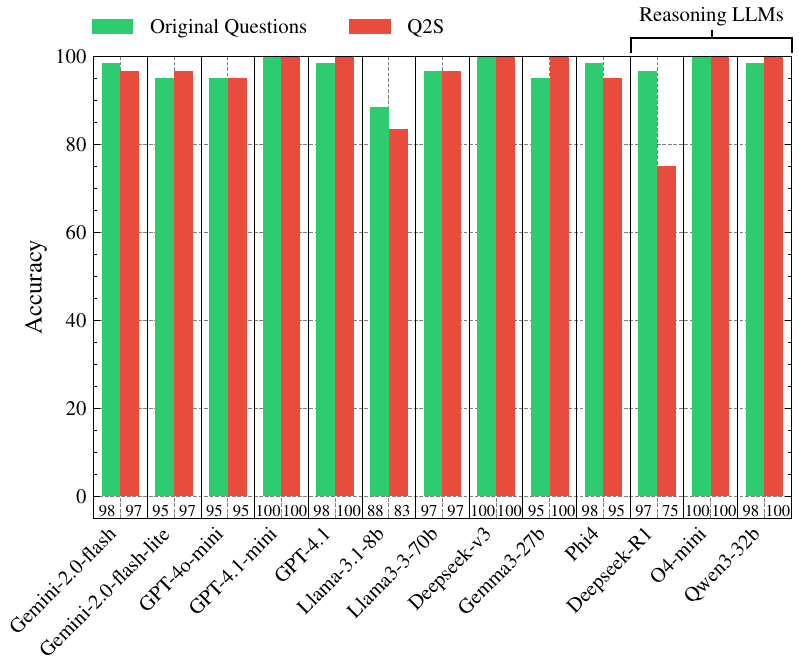}
        \caption{Original vs. Q2S in GSM8K}
        \label{fig:q2s_GSM8K}
    \end{subfigure}
    \hfill
    \begin{subfigure}[b]{0.46\textwidth}
        \centering
        \includegraphics[width=\textwidth]{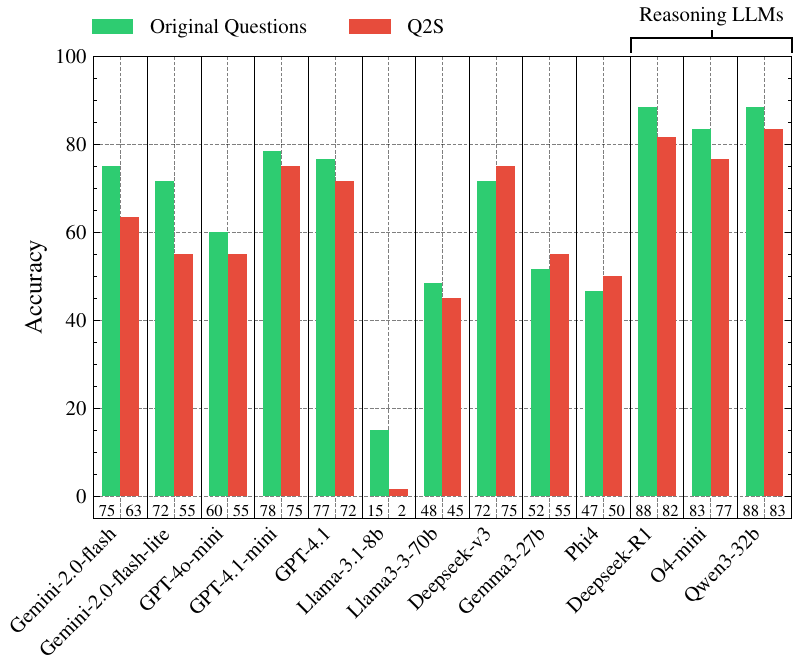}
        \caption{Original vs. Q2S in Physics}     \label{fig:q2s_Physics}
    \end{subfigure}
    
    \caption{Evaluation of different LLMs in original and scenario-based questions}
    \label{fig:q2s}
\end{figure*}

\begin{figure*}[htbp]
    \centering
    \begin{subfigure}[b]{0.46\textwidth}
        \centering
        \includegraphics[width=\textwidth]{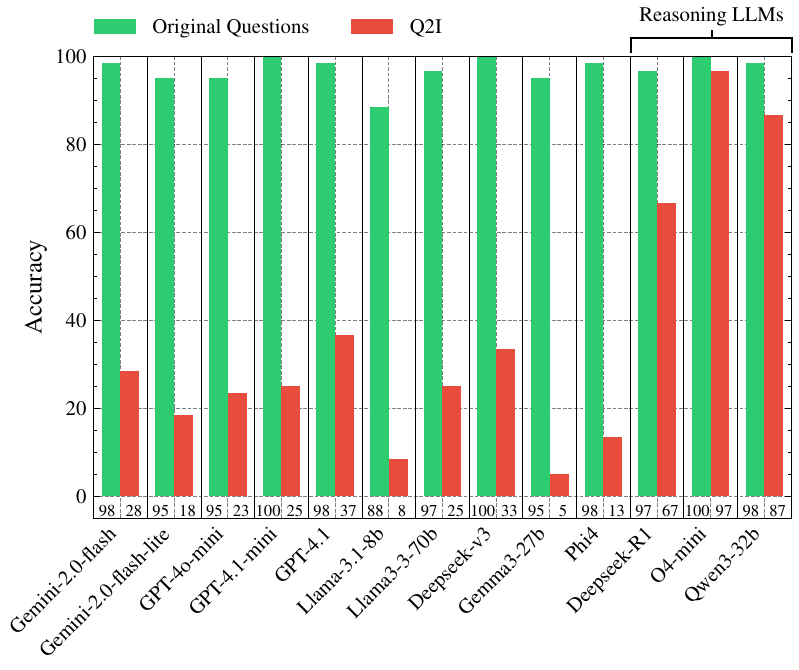}
        \caption{Original vs. Q2I in GSM8K}
        \label{fig:q2i_gsm8k}
    \end{subfigure}
    \hfill
    \begin{subfigure}[b]{0.46\textwidth}
        \centering
        \includegraphics[width=\textwidth]{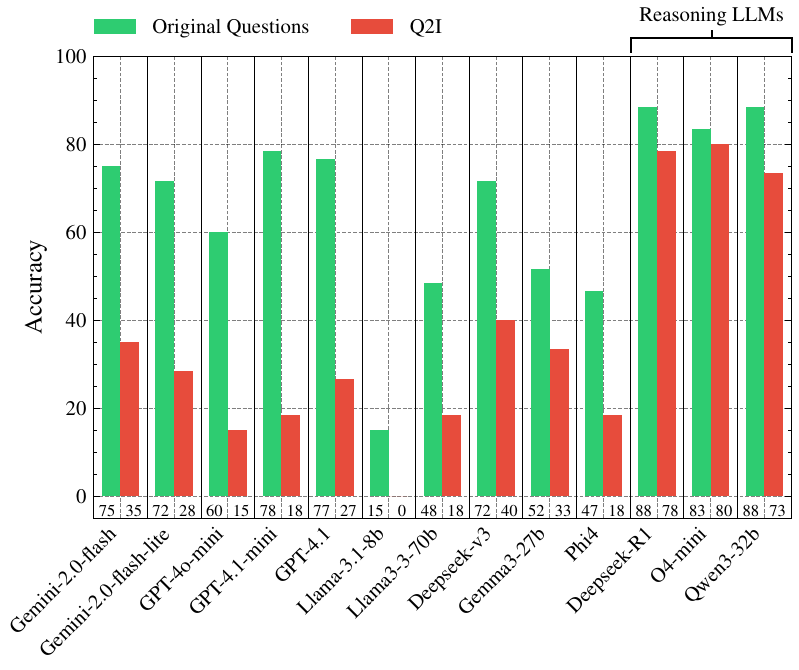}
        \caption{Original vs. Q2I in Physics}
        \label{fig:q2i_physics}
    \end{subfigure}
    
    \caption{Evaluation of different LLMs in original and instruction-based questions}
    \label{fig:q2i}
\end{figure*}

\section{Experiment Setup}
\label{sec:setup}

We constructed the \textsc{DeepQuestion} benchmark by applying our framework to 60 randomly selected GSM8K questions (in English) and 60 physics questions from the Iranian University Entrance Exam (in Persian).
While our method aims to minimize reliance on expert intervention, these datasets were chosen due to the familiarity of the authors to math and physics subjects. The generated deep questions were manually reviewed for quality and accuracy, while we evaluated the framework by comparing them to the original questions.

\begin{figure*}[!t]
    \centering
    \begin{subfigure}[b]{0.46\textwidth}
        \centering
        \includegraphics[width=\textwidth]{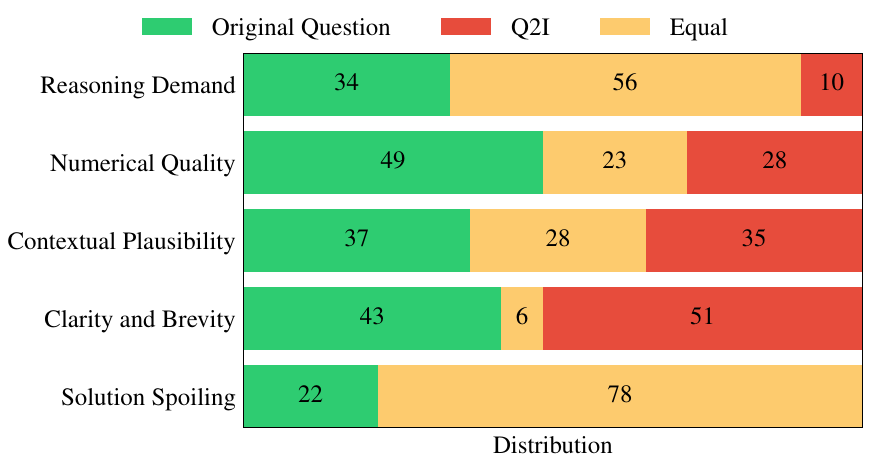}
        \caption{GSM8K}
        \label{fig:win_o4_gsm8k}
    \end{subfigure}
    \hfill
    \begin{subfigure}[b]{0.46\textwidth}
        \centering
        \includegraphics[width=\textwidth]{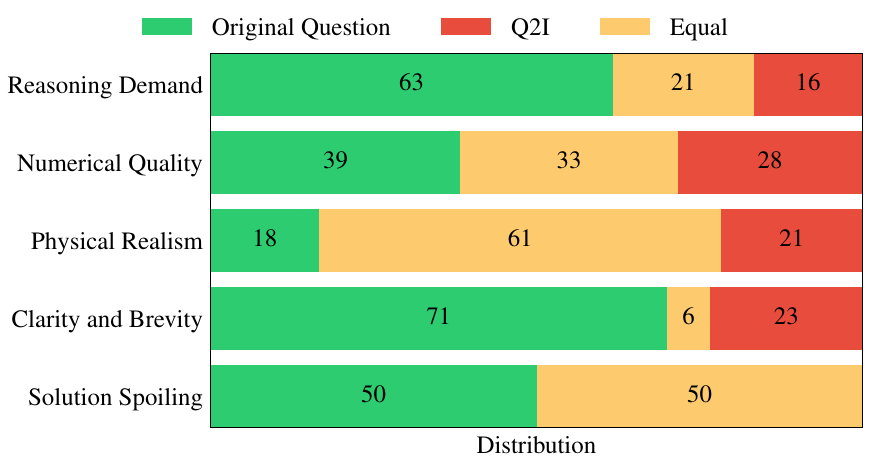}
        \caption{Physics}
        \label{fig:win_o4_physics}
    \end{subfigure}
    
    \caption{Win rate of original against Q2I questions for O4-mini}
    \label{fig:o4_quality}
\end{figure*}

\begin{figure*}[!t]
    \centering
    \begin{subfigure}[b]{0.46\textwidth}
        \centering
        \includegraphics[width=\textwidth]{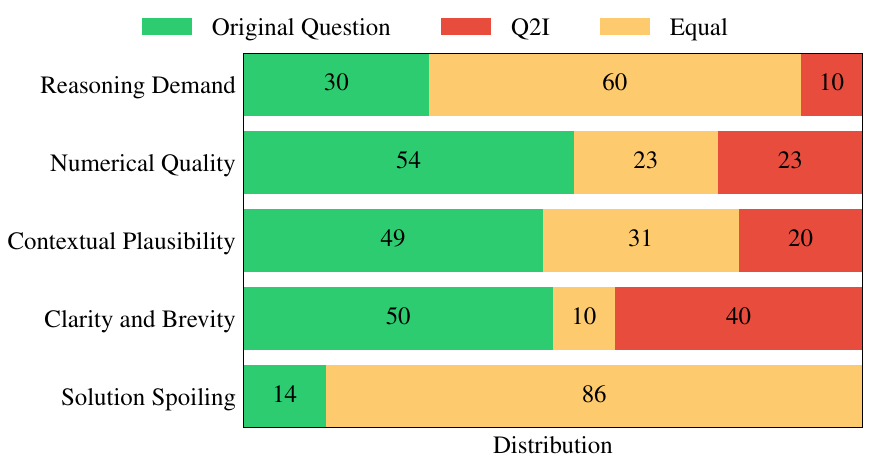}
        \caption{GSM8K}
        \label{fig:win_deepseek_gsm8k}
    \end{subfigure}
    \hfill
    \begin{subfigure}[b]{0.46\textwidth}
        \centering
        \includegraphics[width=\textwidth]{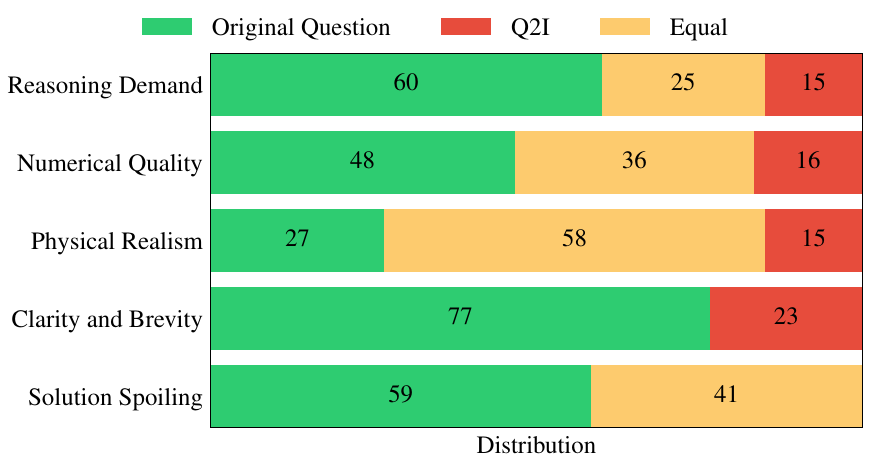}
        \caption{Physics}
        \label{fig:win_deepseek_physics}
    \end{subfigure}
    
    \caption{Win rate of original against Q2I questions for Deepseek-R1}
    \label{fig:deepseek_quality}
\end{figure*}

\begin{figure*}[!t]
    \centering
    \begin{subfigure}[b]{0.46\textwidth}
        \centering
        \includegraphics[width=\textwidth]{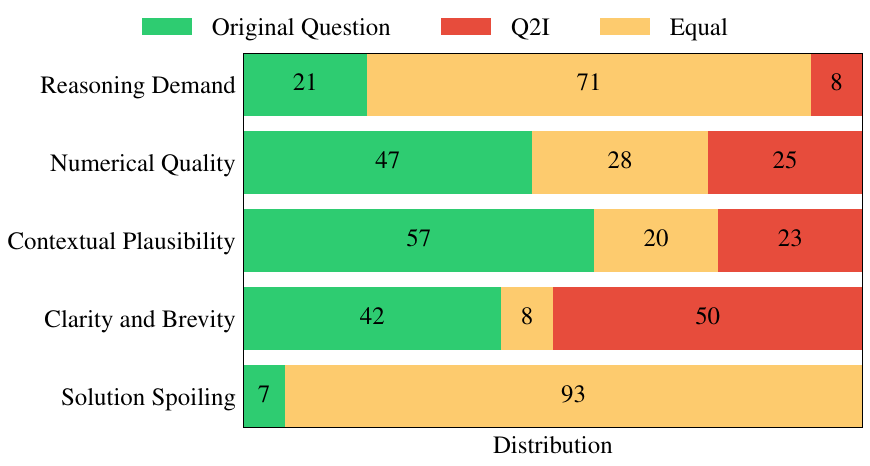}
        \caption{GSM8K}
        \label{fig:win_qwen_gsm8k}
    \end{subfigure}
    \hfill
    \begin{subfigure}[b]{0.46\textwidth}
        \centering
        \includegraphics[width=\textwidth]{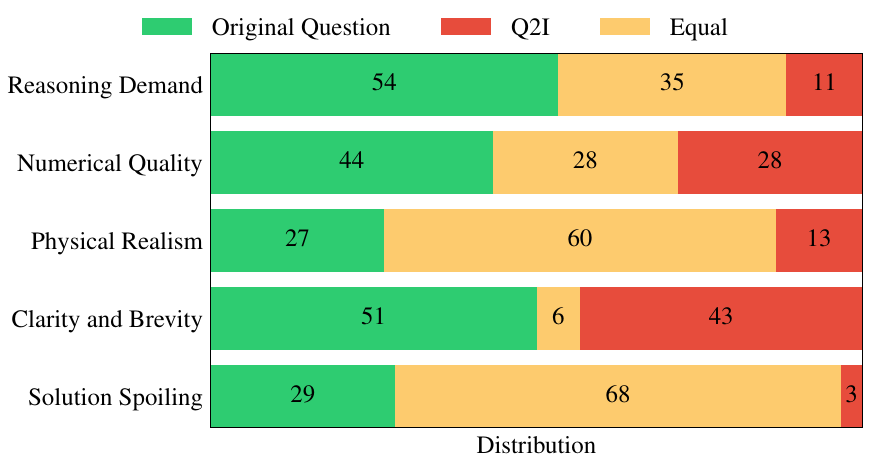}
        \caption{Physics}
        \label{fig:win_qwen_physics}
    \end{subfigure}
    
    \caption{Win rate of original against Q2I questions for Qwen3-32b}
    \label{fig:qwen3_quality}
\end{figure*}

For evaluation, we utilized Gemini-2, GPT-4.1, Llama-3.1~\cite{grattafiori2024llama3}, Deepseek-V3~\cite{liu2024deepseekV3}, Deepseek-R1~\cite{guo2025deepseekR1}, O4-mini, Gemma3~\cite{team2025gemma3}, Phi4~\cite{abdin2024phi4}, and Qwen3~\cite{yang2025qwen3}.
To ensure reproducibility, all models were run with a temperature setting of zero.
Furthermore, within the framework itself, we employed Gemini 2.5 Pro in prompt generator LLM, prompt evaluator LLM, and question generator LLM.

In the Q2S setting, the model generates a solution to a given scenario-question, and correctness is easily verified due to well-defined answers in math and physics problems in our setup.

In the Q2I setting, the model generates questions from instructions. Evaluation consisted of two steps: (1) Answerability Check: verifying whether a strong model (O4-mini) and the model itself can correctly solve the generated questions, indicating the validity of the questions. (2) LLM-as-Judge: since answerability alone does not capture question quality, a powerful language model (O4-mini) was employed for direct qualitative evaluation based on expert-defined criteria: Reasoning Demand — how much genuine thinking and decision-making the question requires; Numerical Quality — whether the numerical values are realistic and meaningful; Physical Realism — whether the scenario is plausible and internally consistent; Clarity and Brevity — how concise and understandable the question is; and Solution Spoiling — whether the problem avoids revealing its own answer or solution steps.

In future work, for open-ended questions that do not have a single correct answer, the same LLM-as-Judge approach could be extended. By defining domain-specific evaluation criteria, LLMs can provide consistent, qualitative assessments that capture nuanced aspects of question quality beyond simple answerability.

\begin{table*}[t]
\centering
\caption{Model performance across translation tasks.}
\label{tab:combined}

\begin{subtable}[t]{\textwidth}
\centering
\caption{Physics – Translate to English}
\label{sub_tab:physics}
\resizebox{\textwidth}{!}{
\begin{tabular}{lcccccc}
\toprule
Models & Llama3-3-70b & Llama3-3-70b-Trans & GPT-4.1 & GPT-4.1-Trans & O4-mini & O4-mini-Trans \\
\midrule
Original Questions & 48.33\% & 61.67\% & 76.67\% & 78.33\% & 83.33\% & 78.33\% \\
Q2S                & 45.00\% & 50.00\% & 71.67\% & 75.00\% & 76.67\% & 80.00\% \\
Q2I                & 18.33\% & 23.33\% & 26.67\% & 33.33\% & 80.00\% & 73.33\% \\
\bottomrule
\end{tabular}}
\end{subtable}

\vspace{0.4cm}

\begin{subtable}[t]{\textwidth}
\centering
\caption{GSM8K – Translate to Persian}
\label{sub_tab:gsm8k}
\resizebox{\textwidth}{!}{
\begin{tabular}{lcccccc}
\toprule
Models & Llama3-3-70b & Llama3-3-70b-Trans & GPT-4.1 & GPT-4.1-Trans & O4-mini & O4-mini-Trans \\
\midrule
Original Questions & 96.67\% & 88.33\% & 98.33\% & 96.67\% & 100.00\% & 100.00\% \\
Q2S                & 96.67\% & 75.00\% & 100.00\% & 95.00\% & 100.00\% & 98.33\% \\
Q2I                & 25.00\% & 11.67\% & 36.67\% & 31.67\% & 96.67\% & 95.00\% \\
\bottomrule
\end{tabular}}
\end{subtable}

\end{table*}

\section{Results}
\label{sec:result}

Based on the distinctions between general-purpose models and reasoning-focused models, we evaluated LLMs from both categories and report our results. Our evaluation methodology is similar to that of Illusion of Thinking, which also compared these two families of models.
\subsection{General-Purpose LLMs}
First, we evaluated general-purpose LLMs on question generation using Q2S.
As shown in \Cref{fig:q2s}, general-purpose models exhibit a moderate decrease in performance on Q2S questions. Some models, such as Gemini-2-Flash and LLaMA-3.1-8B, demonstrate a noticeable reduction in accuracy, whereas others, including Gemma-3-27B and DeepSeek-v3, achieve comparable or even improved performance on Q2S questions. These findings indicate that general-purpose LLMs can solve scenario-based questions with only a slight decrease in accuracy, suggesting their capability to apply knowledge at the applying level of Bloom’s taxonomy.

To further assess LLMs’ abilities at higher levels, such as the creation level, we were motivated to design the Q2I experiment. In this setting, models exhibited a substantial drop in accuracy when tackling tasks that require deeper understanding and more complex reasoning.
No model surpassed 38\% accuracy on instruction-based question generation for GSM8K or physics (see \Cref{fig:q2i}), despite over 95\% on original questions. These results support the use of Bloom’s taxonomy as an evaluation framework and reveal the limitations of general-purpose LLMs on higher-order cognitive tasks.

\subsection{Reasoning LLMs}
Based on the observed results from general-purpose models and their performance decline on Q2I questions, we evaluated reasoning models on both Q2I and Q2S questions. Reasoning models exhibited a performance trend similar to that of general-purpose models on Q2S questions. However, for Q2I questions, the performance reduction in reasoning models was smaller. Despite this, accuracy in the Q2I setting still decreased by 8–30\% (\Cref{fig:q2i}), highlighting limitations of these models at higher cognitive levels. For example, Deepseek-R1’s performance declined by 30\% on GSM8K and 10\% on physics datasets, whereas Qwen-3-32B experienced an 11\% drop on both datasets. Although reasoning models generally outperform general-purpose models, significant gaps remain in achieving deep and comprehensive understanding. These findings indicate that neither general-purpose nor reasoning models achieve strong performance at the creation level of Bloom’s taxonomy, suggesting that current LLMs are still distant from high-level cognitive reasoning.

\subsection{Question Quality Analysis}
While reasoning models can generate questions that follow instructions and maintain topical consistency, it is crucial to assess their quality beyond surface-level compliance. To this end, we use a preference-based evaluation with O4-mini as a judge, directly comparing generated and original questions across five criteria: reasoning demand, numerical quality, physical realism, clarity and brevity, and solution spoiling. This approach addresses concerns about the model’s ability to assess quality by focusing on direct comparison.

As illustrated in \Cref{fig:o4_quality,fig:deepseek_quality,fig:qwen3_quality}, the physics questions generated by the models continue to underperform compared to the original questions. Specifically, in the dimensions of Reasoning Demand, Clarity, and Brevity, the original questions outperform the generated ones in more than 50\% of cases across all three models. Furthermore, in terms of Solution Spoiling, Q2I exhibits negligible improvement, achieving wins in only approximately 3\% of cases. These findings underscore the persistent challenges faced by reasoning models in producing high-quality questions. In contrast, generated GSM8K questions are more comparable to the originals, likely reflecting the relative simplicity of this dataset. Overall, the results demonstrate the effectiveness of our framework in enhancing question depth without introducing artificial difficulty. Nevertheless, a substantial quality gap remains, emphasizing the ongoing difficulty of generating high-quality questions even with advanced reasoning models.

\subsection{Disentangling Linguistic Effects from Cognitive Complexity}
To rule out linguistic artifacts, we conducted a cross-lingual experiment by translating the Persian Physics dataset into English (see \Cref{sub_tab:physics}) and the English GSM8K dataset into Persian (see \Cref{sub_tab:gsm8k}).
We evaluated three representative models, including Llama3-70B, GPT-4.1, and O4-mini, to test whether performance declines in  \textsc{DeepQuestion} stem from increased cognitive demand or from language-related confounds.

Across both datasets and translation directions, the performance hierarchy (Original > Q2S > Q2I) remained consistent.
For instance, in the translated Physics dataset, O4-mini scored 83.33\% (Original), 80.00\% (Q2S), and 73.33\% (Q2I); GPT-4.1 followed the same pattern (78.33\% → 75.00\% → 33.33\%). 
Similarly, in Persian-translated GSM8K, Llama3-70B dropped from 88.33\% to 75.00\% and then to 11.67\%.
This stability confirms that our transformations genuinely elevate cognitive complexity independent of language.

These results confirm that \textsc{DeepQuestion} isolates cognitive complexity as defined by Bloom’s Taxonomy.
Observed declines arise from reasoning demands rather than language artifacts, validating our framework’s robustness across languages and model families.

\section{Conclusion}
\label{sec:conclusion}
This work presented \textsc{DeepQuestion}, a systematic and scalable framework for generating cognitively diverse benchmarks that extend existing datasets through the lens of Bloom’s taxonomy. By transforming conventional question–answer pairs into scenario-based and instruction-driven tasks, \textsc{DeepQuestion} provides a structured means to probe LLMs across multiple levels of cognition, from application to creation.

Our findings across state-of-the-art LLMs reveal a persistent decline in performance as task complexity and cognitive depth increase—reaching up to 70\% accuracy loss in higher-order reasoning tasks. These results highlight that current LLMs, despite their success on standardized benchmarks, still exhibit shallow generalization and limited conceptual transfer when confronted with real-world or creative problem-solving scenarios.

Beyond performance assessment, \textsc{DeepQuestion} offers a replicable pathway for future benchmark development. The framework can be readily applied to other domains such as law, medicine, or engineering to examine domain-specific reasoning. Moreover, it opens new directions for exploring automated benchmark construction, self-improving evaluation pipelines, and curriculum-style training data aligned with cognitive learning theories.

In summary, \textsc{DeepQuestion} bridges educational psychology and AI evaluation, revealing fundamental gaps in LLM reasoning and offering tools to measure and close them. As language models continue to evolve, cognitively grounded and contextually rich evaluation frameworks like \textsc{DeepQuestion} will be essential for steering their progress toward genuine understanding and human-aligned intelligence.

\newpage
\section{Bibliographical References}\label{sec:reference}

\bibliographystyle{lrec2026-natbib}
\bibliography{lrec2026-example}

\bibliographystylelanguageresource{lrec2026-natbib}
\bibliographylanguageresource{languageresource}

\end{document}